\documentclass[sigconf,nonacm]{acmart}

\usepackage{algorithm}
\usepackage{algpseudocode}
\usepackage{enumitem}
\usepackage{tikz}
\usetikzlibrary{arrows.meta,calc}
\definecolor{figink}{HTML}{27333D}
\definecolor{figmuted}{HTML}{58636D}
\definecolor{figline}{HTML}{C4CDD4}
\definecolor{figpurple}{HTML}{54738A}
\definecolor{figpurplebg}{HTML}{F0F4F7}
\definecolor{figorange}{HTML}{9B8874}
\definecolor{figorangebg}{HTML}{FAF7F3}
\definecolor{figteal}{HTML}{768692}
\definecolor{figtealbg}{HTML}{F6F8F9}
\definecolor{figgraybg}{HTML}{F6F7F8}
\tikzset{
  paper diagram/.style={font=\sffamily\fontsize{9}{11}\selectfont,
    text=figink, line width=.60pt, >={Stealth[length=1.7mm,width=1.2mm]}},
  diagram box/.style={draw=figteal,fill=white,rounded corners=1pt,
    align=center,inner sep=4pt,minimum height=1cm},
  proposal box/.style={diagram box,draw=figpurple,fill=figpurplebg},
  validation box/.style={diagram box,draw=figorange,fill=figorangebg},
  pool box/.style={diagram box,draw=figteal,fill=figtealbg},
  diagram arrow/.style={->,draw=figmuted,line width=.65pt},
  diagram note/.style={font=\sffamily\fontsize{8}{10}\selectfont,text=figmuted,align=center},
  diagram heading/.style={font=\sffamily\bfseries\fontsize{9}{11}\selectfont,anchor=west},
  diagram small/.style={font=\sffamily\fontsize{8}{10}\selectfont,align=center}
}

\setcopyright{none}
\title[Explore More, Validate Less]{Explore More, Validate Less: Efficient Reflective Prompt Optimization}
\author{Xiaofan Zhou}
\affiliation{\institution{Pennsylvania State University}\country{USA}}
\email{xpz5429@psu.edu}
\author{Lu Cheng}
\affiliation{\institution{Pennsylvania State University}\country{USA}}

\begin{document}

\begin{abstract}
Reflective prompt optimization improves instructions using task feedback, but coupling candidate generation with full-validation-based selection can make search costly and sequential. We propose a two-stage framework that decouples reflective candidate-pool construction from validation allocation. During construction, sampled parents first answer training questions, and groups of their saved responses guide parallel reflection and revision. Each child is screened on its group's questions against the corresponding parent scores. Improving, nonduplicate prompts enter the pool and can become parents without full validation. After construction, an exploration policy selects candidates from the frozen pool under a fixed full-validation budget, and the prompt with the highest measured validation score is returned. This separation allows candidate-pool size and validation budget to be set independently. We evaluate random selection and a multilayer perceptron with similarity-based exploration (MLP + similarity) on five benchmarks using Qwen3.5-9B and Llama-3.1-8B-Instruct. With 100 retained candidates and 20 full validations, our framework improves over the initial solver by up to 11.22 percentage points. MLP + similarity improves over the initial solver in all ten comparisons, although neither selection policy consistently outperforms the other. These results support separating reflective search from budgeted validation while leaving room for stronger candidate-selection policies.
\end{abstract}


\maketitle

\section{Introduction}

Large language models (LLMs) exhibit broad capabilities in language and reasoning \citep{openai2023gpt4,anthropic2024claude3,grattafiori2024llama3}, but their task performance depends on how these capabilities are elicited. In-context learning (ICL) specifies tasks through demonstrations without updating model weights \citep{brown2020fewshot}; these demonstrations convey input distributions, label spaces, and output formats \citep{min2022demonstrations}. Chain-of-thought (CoT) prompting improves multi-step reasoning through worked examples of intermediate steps \citep{wei2022cot}, while zero-shot CoT can elicit such steps with a short instruction \citep{kojima2022zeroshot}. Prompt design is also consequential within a task: example selection \citep{liu2022examples}, example order \citep{lu2022ordered}, and prompt format \citep{zhao2021calibrate} can substantially affect performance. These findings make prompts a practical optimization target. Manually inspecting outputs and revising instructions is costly across tasks, motivating automatic prompt optimization that improves prompts while keeping model parameters fixed \citep{zhou2023ape,opsahl-ong-etal-2024-optimizing,agrawal2025gepa}.
For LLM-based Web content analysis and question answering, the relevant challenge is how to adapt task instructions while limiting repeated evaluation costs. We study this trade-off between prompt exploration and validation expenditure.

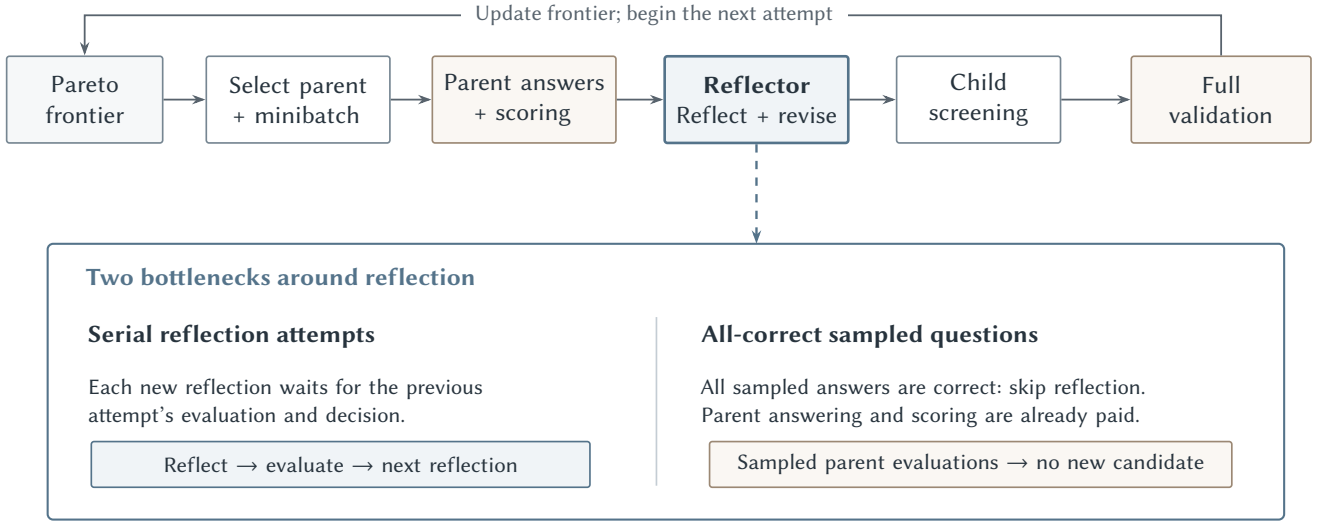
\begin{figure*}[t]
    \centering
    \resizebox{\textwidth}{!}{% Original single-proposal GEPA: one overview, one reflector callout.
\begin{tikzpicture}[paper diagram]
\path[use as bounding box] (0,0) rectangle (17,7.65);
\node[diagram heading] at (.10,7.36) {GEPA: the original single-proposal loop};

% The main pipeline remains visually separate from the explanatory callout.
\node[pool box,text width=1.70cm,minimum height=1.10cm]
  (frontier) at (1.12,5.78) {Pareto\\frontier};
\node[diagram box,text width=2.05cm,minimum height=1.10cm]
  (select) at (3.83,5.78) {Select parent\\+ minibatch};
\node[validation box,text width=2.05cm,minimum height=1.10cm]
  (parent) at (6.70,5.78) {Parent answers\\+ scoring};
\node[proposal box,line width=.95pt,text width=2.05cm,minimum height=1.10cm]
  (reflect) at (9.65,5.78) {\textbf{Reflector}\\Reflect + revise};
\node[diagram box,text width=1.80cm,minimum height=1.10cm]
  (child) at (12.47,5.78) {Child\\screening};
\node[validation box,text width=2.00cm,minimum height=1.10cm]
  (full) at (15.55,5.78) {Full\\validation};
\draw[diagram arrow] (frontier) -- (select);
\draw[diagram arrow] (select) -- (parent);
\draw[diagram arrow] (parent) -- (reflect);
\draw[diagram arrow] (reflect) -- (child);
\draw[diagram arrow] (child) -- (full);

% The return edge explains why successive proposals wait for a decision.
\draw[diagram arrow] (full.north) -- (15.55,6.85) -- (1.12,6.85) -- (frontier.north);
\node[diagram note,fill=white,inner xsep=5pt] at (8.35,6.85)
  {Update frontier; begin the next attempt};

% A single dashed arrow attaches both bottlenecks to the reflector.
\draw[diagram arrow,draw=figpurple,dashed,line width=.85pt]
  (reflect.south) -- (9.65,3.95);
\path[draw=figpurple,fill=white,rounded corners=2pt,line width=.75pt]
  (.65,.45) rectangle (16.35,3.95);
\node[diagram heading,text=figpurple] at (1.02,3.53)
  {Two bottlenecks around reflection};
\draw[draw=figline] (8.38,.88) -- (8.38,3.02);

\node[diagram heading] at (1.05,2.78) {Serial reflection attempts};
\node[diagram small,anchor=north west,align=left,text width=6.60cm] at (1.05,2.35)
  {Each new reflection waits for the previous\\attempt's evaluation and decision.};
\node[proposal box,text width=6.05cm,minimum height=.58cm,
  font=\sffamily\fontsize{8}{10}\selectfont] at (4.37,1.15)
  {Reflect $\rightarrow$ evaluate $\rightarrow$ next reflection};

\node[diagram heading] at (8.83,2.78) {All-correct sampled questions};
\node[diagram small,anchor=north west,align=left,text width=6.95cm] at (8.83,2.35)
  {All sampled answers are correct: skip reflection.\\Parent answering and scoring are already paid.};
\node[validation box,text width=6.35cm,minimum height=.58cm,
  font=\sffamily\fontsize{8}{10}\selectfont] at (12.37,1.15)
  {Sampled parent evaluations $\rightarrow$ no new candidate};
\end{tikzpicture}}
    \Description{The original single-proposal GEPA pipeline: select a parent from the Pareto frontier, sample questions, obtain parent answers and scores, reflect and revise when errors are present, screen the child, and fully validate an improved child before updating the frontier. A dashed annotation arrow from the reflector points to two bottlenecks: serial reflection attempts, and parent evaluation costs already incurred when an all-correct minibatch causes reflection to be skipped.}
    \caption{\textbf{GEPA's original single-proposal workflow.} The reflector callout highlights sequential evaluation dependencies and parent-evaluation cost when all sampled answers are correct. Solid arrows show the successful-proposal path and frontier update; return edges for all-correct minibatches and rejected children are omitted for clarity. The dashed arrow connects the reflector to explanatory annotations rather than an execution step.}
    \label{fig:motivation}
\end{figure*}

Reflective prompt optimization faces several challenges. First, reflection attempts can consume time without producing an improved prompt. Each attempt requires parent evaluation and, when errors are found, reflection, revision, and child screening; these costs are incurred even when no candidate is retained. For example, Genetic-Pareto (GEPA) retains complementary parents, but its original single-proposal loop evaluates a parent on a training minibatch, revises it using observed errors, and fully validates a locally improved child before updating the frontier \citep{agrawal2025gepa}. As Figure~\ref{fig:motivation} illustrates, successive reflections depend on preceding evaluations and decisions. An all-correct minibatch incurs parent-answering and scoring costs but produces no revision; a rejected child also consumes work without expanding the pool. Additionally, repeated full validation during prompt evolution can be time-consuming, limiting the range of prompts explored within a fixed time budget. Parallel proposals reduce serial generation, yet accepted proposals still require full validation before frontier updates \citep{zhang2026parallel}.

To address the above challenges, our approach separates reflective candidate-pool construction from validation-based selection, exploring multiple lineages before fully validating a subset of candidates. We first evaluate each sampled parent on a set of training questions, then reuse small groups of its saved responses, each containing an observed error, for parallel reflection and revision. Children that outperform the saved parent scores on their corresponding groups enter the pool after deduplication and become eligible for later expansion. Once the pool is frozen, an exploration policy, such as random sampling or MLP + similarity, selects candidates for full validation under a fixed budget. The candidate with the highest measured validation score is returned. Removing full validation from the generation loop allows the pool to grow beyond the validation budget, while different selection policies can reuse the same pool without regenerating prompts.

Our contributions are threefold:
\begin{itemize}[leftmargin=1.3em,labelsep=0.4em,topsep=4pt,itemsep=3pt,parsep=0pt,partopsep=0pt]
    \item We introduce a two-stage framework that decouples reflective candidate-pool construction from full validation, allowing pool size and validation budget to be set independently. This targets validation efficiency by reserving expensive full evaluations for a selected subset rather than every accepted revision.
    \item We develop a multi-parent construction procedure that shares training feedback across parallel reflection branches and locally screens revisions without full-validation scores. Reusing parent responses reduces repeated evaluation, while concurrent expansion across parents and minibatches supports broader search over alternative revision paths.
    \item Experiments on five benchmarks with two models demonstrate the effectiveness of our framework, with gains over the initial solver of up to 11.22 percentage points. Our variants use 20 full validations versus GEPA's 30 and collectively lead on seven of ten model--dataset combinations. Separate analyses report stage-wise runtime and show greater mean tree depth in nine of ten comparisons against the configured GEPA searches.
\end{itemize}

\section{Related Work}

\subsection{Automatic Prompt Optimization}
Automatic prompt optimization searches for effective prompts while keeping model weights fixed. AutoPrompt uses gradient-guided search to find discrete trigger tokens for masked language models \citep{shin2020autoprompt}. For natural-language instructions, APE generates candidates from demonstrations \citep{zhou2023ape}; Optimization by Prompting (OPRO) conditions proposals on past instructions and scores \citep{yang2024opro}; EvoPrompt evolves prompt populations \citep{guo2024evoprompt}. MIPRO jointly optimizes instructions and demonstrations for multi-stage programs \citep{opsahl-ong-etal-2024-optimizing}. Beyond individual prompts, AFlow searches code-represented agentic workflows using Monte Carlo tree search and execution feedback \citep{zhang2025aflow}. Consensus-Evolve (C-Evolve) evolves prompt groups using voting-based fitness rather than individual scores \citep{li2025cevolve}. metaTextGrad optimizes the prompts and structure of language-model optimizers themselves \citep{xu2025metatextgrad}.

Evaluation-efficient selection is also established. Prompt Optimization with Textual Gradients (ProTeGi) combines textual feedback, beam search, and bandit-based selection \citep{pryzant2023protegi}. InstructZero uses Bayesian optimization over low-dimensional soft prompts that an open model converts into instructions for a black-box model \citep{chen2024instructzero}. Instruction Optimization using Neural Bandits Coupled with Transformers (INSTINCT) uses a multilayer perceptron (MLP) surrogate coupled to transformer representations to guide soft-prompt search and instruction generation \citep{lin2024instinct}. Multi-Agent System Prompt Optimization via Bandits (MASPOB) combines graph neural representations, uncertainty-aware selection, and coordinate ascent to optimize prompts within fixed multi-agent workflows \citep{hong2026maspob}.

TRIPLE formulates selection from a pre-generated prompt pool as fixed-budget best-arm identification \citep{shi2024triple}. Building on budgeted selection, our focus is constructing a reflective candidate tree whose intermediate prompts can produce descendants without full validation, using shared parent feedback and local child screening. The subsequent policy allocates complete validation-set evaluations over that frozen pool. Unlike soft-prompt search, it selects among already constructed natural-language prompts; random selection and MLP + similarity provide two ways to explore the same frozen pool.

\subsection{Feedback-driven Improvement}
Self-Refine critiques and revises individual outputs \citep{madaan2023selfrefine}, while Reflexion stores feedback-derived reflections for subsequent agent trials \citep{shinn2023reflexion}. TextGrad propagates language-model feedback through a computation graph to optimize its components \citep{yuksekgonul2025textgrad}. This addresses how feedback reaches optimization variables, whereas our framework controls when candidate prompts receive full validation. GEPA combines trajectory reflection with instance-wise Pareto selection to retain complementary prompts \citep{agrawal2025gepa}. Its parallel-proposal extension generates revisions from multiple parents, but still fully validates accepted proposals before updating the frontier \citep{zhang2026parallel}. In our framework, training feedback and local screening drive pool construction, allowing accepted children to become parents before full validation. Validation is allocated after the pool is frozen. The optimization target remains a reusable task prompt rather than an individual output or the optimizer itself.

\section{Method}
\label{sec:method}

\subsection{Problem Definition}
\label{sec:problem_definition}
We seek a reusable task prompt that performs well on unseen examples under a limited evaluation budget. Let $f_p$ denote an LLM conditioned on prompt $p$, with model weights and the task interface held fixed. For an input $x$ and target $y$, a task-specific scoring function $s(f_p(x),y)\in[0,1]$ measures output quality. Optimization uses a training partition $\mathcal{D}_{\mathrm{tr}}$ for feedback and a validation partition $\mathcal{D}_{\mathrm{val}}$ to measure the full-validation score $V(p)$:
\begin{equation}
\label{eq:validation_score}
    V(p)=\frac{1}{|\mathcal{D}_{\mathrm{val}}|}
    \sum_{(x,y)\in\mathcal{D}_{\mathrm{val}}}s(f_p(x),y).
\end{equation}
Vertical bars denote set cardinality or minibatch length. We distinguish the candidate-pool size $N$ from the full-validation budget $B$, both positive integers with $B\leq N$. One budget unit evaluates one candidate on all of $\mathcal{D}_{\mathrm{val}}$, rather than on a single example. Given an initial prompt $p_0$, the problem is to construct a pool $\mathcal{P}$ of size $N$, choose an evaluated subset $\mathcal{O}\subseteq\mathcal{P}$ with $|\mathcal{O}|\leq B$, and return a prompt from $\mathcal{O}$. Training evaluations and reflection calls are accounted for separately from the full-validation budget. Generalization is assessed on held-out test data after final prompt selection.

\subsection{Overall Framework}
\label{sec:framework}
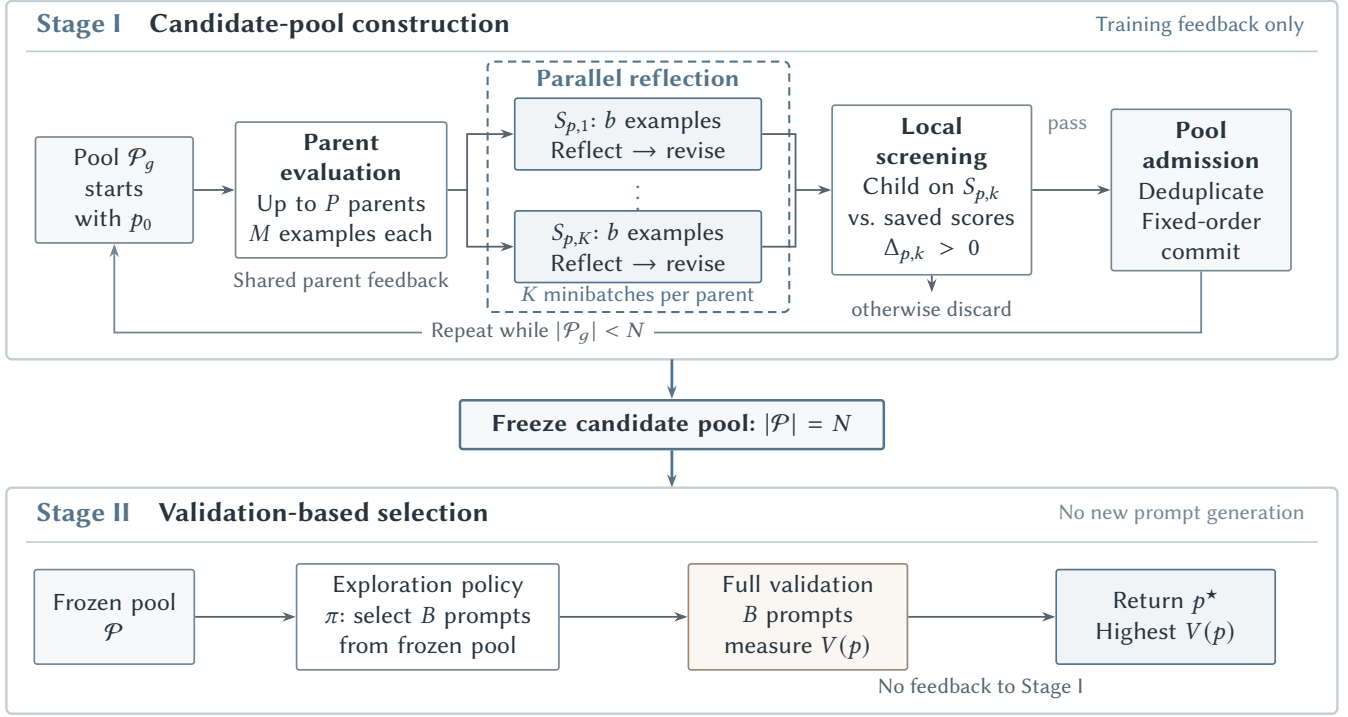
\begin{figure*}[t]
    \centering
    \resizebox{\textwidth}{!}{% Two separate stages, linked only through the frozen candidate pool.
\begin{tikzpicture}[paper diagram]
\path[use as bounding box] (0,0) rectangle (17,9.65);

% Stage boundaries remain visible in grayscale; color is only an accent.
\draw[draw=figline,line width=.85pt,rounded corners=2pt]
  (.05,4.95) rectangle (16.95,9.50);
\node[diagram heading,font=\sffamily\bfseries\fontsize{10}{12}\selectfont]
  at (.32,9.18) {\textcolor{figpurple}{Stage I}\quad Candidate-pool construction};
\node[diagram small,anchor=east,text=figpurple] at (16.65,9.18) {Training feedback only};
\draw[draw=figline] (.30,8.86) -- (16.70,8.86);

\node[pool box,text width=1.72cm,minimum height=1.1cm] (pool) at (1.42,7.10)
  {Pool $\mathcal{P}_g$\\starts with $p_0$};
\node[diagram box,text width=2.40cm,minimum height=1.5cm] (parents) at (4.30,7.10)
  {\textbf{Parent}\\\textbf{evaluation}\\Up to $P$ parents\\$M$ examples each};

% Independent proposal branches are grouped as one parallel execution region.
\draw[draw=figpurple,densely dashed,line width=.75pt,rounded corners=2pt]
  (6.16,5.55) rectangle (9.99,8.73);
\node[diagram small,text=figpurple,font=\sffamily\bfseries\fontsize{9}{11}\selectfont]
  at (8.07,8.53) {Parallel reflection};
\node[proposal box,text width=2.85cm,minimum height=.91cm] (proposal1) at (8.07,7.81)
  {$S_{p,1}$: $b$ examples\\Reflect $\rightarrow$ revise};
\node[proposal box,text width=2.85cm,minimum height=.91cm] (proposal2) at (8.07,6.38)
  {$S_{p,K}$: $b$ examples\\Reflect $\rightarrow$ revise};
\node[text=figpurple] at (8.07,7.11) {$\vdots$};
\node[diagram small,text=figpurple] at (8.07,5.76) {$K$ minibatches per parent};
\node[diagram box,text width=2.25cm,minimum height=1.23cm] (gate) at (11.80,7.10)
  {\textbf{Local screening}\\Child on $S_{p,k}$\\vs.\ saved scores\\$\Delta_{p,k}>0$};
\node[pool box,text width=2.00cm,minimum height=1.23cm] (commit) at (15.22,7.10)
  {\textbf{Pool}\\\textbf{admission}\\Deduplicate\\Fixed-order\\commit};

\draw[diagram arrow] (pool) -- (parents);
\draw[draw=figmuted,line width=.65pt] (parents.east) -- (5.90,7.10);
\draw[diagram arrow] (5.90,7.10) |- (proposal1.west);
\draw[diagram arrow] (5.90,7.10) |- (proposal2.west);
\draw[draw=figmuted,line width=.65pt] (proposal1.east) -- (10.07,7.81) -- (10.07,7.10);
\draw[draw=figmuted,line width=.65pt] (proposal2.east) -- (10.07,6.38) -- (10.07,7.10);
\draw[diagram arrow] (10.07,7.10) -- (gate.west);
\draw[diagram arrow] (gate) -- (commit);
\node[diagram small,text=figteal] at (13.52,7.91) {pass};
\draw[diagram arrow,draw=figmuted] (gate.south) -- (11.80,5.80);
\node[diagram note] at (11.80,5.59) {otherwise discard};

% This feedback loop is entirely inside Stage I.
\draw[diagram arrow,draw=figteal] (commit.south) -- (15.22,5.29) -- (1.42,5.29) -- (pool.south);
\node[diagram small,text=figmuted,fill=white,inner sep=3pt] at (6.80,5.29)
  {Repeat while $|\mathcal{P}_g|<N$};
\node[diagram note] at (4.30,5.93) {Shared parent feedback};

% A single one-way handoff, distinct from the generation loop.
\node[pool box,draw=figpurple,line width=.9pt,text width=5.1cm,
      minimum height=.62cm,font=\sffamily\bfseries\fontsize{9}{11}\selectfont]
  (handoff) at (8.50,4.12) {Freeze candidate pool: $|\mathcal{P}|=N$};
\draw[diagram arrow,draw=figpurple,line width=1pt] (8.50,4.95) -- (handoff.north);
\draw[diagram arrow,draw=figpurple,line width=1pt] (handoff.south) -- (8.50,3.32);

\draw[draw=figline,line width=.85pt,rounded corners=2pt]
  (.05,.45) rectangle (16.95,3.32);
\node[diagram heading,font=\sffamily\bfseries\fontsize{10}{12}\selectfont]
  at (.32,2.98) {\textcolor{figpurple}{Stage II}\quad Validation-based selection};
\node[diagram small,anchor=east,text=figteal] at (16.65,2.98) {No new prompt generation};
\draw[draw=figline] (.30,2.63) -- (16.70,2.63);

\node[pool box,text width=1.75cm,minimum height=1.2cm] (freeze) at (1.42,1.68)
  {Frozen pool\\$\mathcal{P}$};
\node[diagram box,text width=3.05cm,minimum height=1.2cm] (policy) at (5.40,1.68)
  {Exploration policy\\$\pi$: select $B$ prompts\\from frozen pool};
\node[validation box,text width=2.50cm,minimum height=1.2cm] (validate) at (10.10,1.68)
  {Full validation\\$B$ prompts\\measure $V(p)$};
\node[pool box,draw=figpurple,fill=figpurplebg,text width=2.50cm,minimum height=1.2cm] (winner) at (14.77,1.68)
  {Return $p^{\star}$\\Highest $V(p)$};

\draw[diagram arrow] (freeze) -- (policy);
\draw[diagram arrow] (policy) -- (validate);
\draw[diagram arrow] (validate) -- (winner);
\node[diagram note] at (12.42,.77) {No feedback to Stage I};
\node[diagram note,anchor=west] at (.1,.05) {Test evaluation follows prompt selection.};
\end{tikzpicture}}
    \Description{A two-stage workflow. Stage I repeatedly samples parents from a growing pool, shares their saved training responses across parallel reflection branches, and admits locally improved, nonduplicate children. After the pool is frozen, Stage II uses an exploration policy to select candidates for full validation and returns the candidate with the highest measured validation score. There is no feedback from Stage II to Stage I.}
    \caption{\textbf{Two stages separated by a frozen candidate pool.} Stage I constructs $N$ prompts using training feedback only. The dashed region highlights parallel proposal branches: reflection and revision can execute concurrently across minibatches and parents, while each revision follows its own reflection. Locally improved children enter the pool, and the generation loop stays within Stage I. Once the pool reaches $N$ prompts, its texts and metadata are frozen. Stage II uses an exploration policy $\pi$ to allocate $B$ full validations and returns the prompt with the highest measured validation score, without generating new prompts or sending validation feedback to Stage I. Construction and selection symbols are defined in Sections~\ref{sec:pool_generation} and~\ref{sec:validation_selection}, respectively. Test evaluation follows final selection.}
    \label{fig:framework}
\end{figure*}

Our framework separates candidate-pool construction from vali\-dation-based selection. The first stage follows four steps: parent evaluation, parallel reflection, local screening, and pool admission. Multiple parents supply training feedback for parallel revisions, and locally improved children enter the pool. At the stage boundary, prompt texts and generation metadata are frozen. The second stage applies an exploration policy to this pool, acquiring validation scores without generating additional candidates. Different exploration policies can therefore share the same construction stage. The final output is the prompt with the highest measured validation score. Figure~\ref{fig:framework} illustrates the framework, and Algorithm~\ref{alg:two_stage} summarizes the procedure.

% Algorithmic summary of Sections 3.3--3.4; use the manuscript's notation.
\begin{algorithm}[t]
    \caption{Two-stage prompt optimization}
    \label{alg:two_stage}
    \small
    \algrenewcommand\algorithmicindent{0.8em}
    \algrenewcommand\algorithmicrequire{\textbf{Input:}}
    \algrenewcommand\algorithmicensure{\textbf{Output:}}
    \begin{algorithmic}[1]
        \Require Initial prompt $p_0$; data $\mathcal{D}_{\mathrm{tr}},\mathcal{D}_{\mathrm{val}}$;
        pool size $N$; validation budget $B$; parent limit $P$;
        feedback size $M$; branches $K$; minibatch size $b$;
        exploration policy $\pi$.
        \Ensure Selected prompt $p^{\star}$.
        \Statex \textbf{Stage I: Candidate-pool construction}
        \State $\mathcal{P}_0\gets\{p_0\}$; $g\gets0$
        \While{$|\mathcal{P}_g|<N$}
            \State Uniformly select up to $P$ distinct parents from $\mathcal{P}_g$
            \ForAll{selected parents $p$ \textbf{in parallel}}
                \Repeat
                    \State Sample $M$ distinct examples $T_p\subseteq\mathcal{D}_{\mathrm{tr}}$
                    \State Evaluate $p$ on $T_p$; save responses, scores, and feedback
                \Until{at least one response fails}
                \State Sample $K$ distinct ordered size-$b$ minibatches $S_{p,1},\ldots,S_{p,K}$ from $T_p$, each containing a failure and no repeated example
                \For{$k=1,\ldots,K$ \textbf{in parallel}}
                    \State $r_{p,k}\sim R(\,\cdot\mid p,\tau(p,S_{p,k}))$
                    \State $p'_{p,k}\sim G(\,\cdot\mid p,r_{p,k})$
                    \State Evaluate $p'_{p,k}$ on $S_{p,k}$; compute $\Delta_{p,k}$ against saved parent scores
                \EndFor
                \State Gather improving children $\mathcal{Q}_p$ using Eq.~\eqref{eq:local_gate}
            \EndFor
            \State Join all branches; merge $\mathcal{Q}_p$ across parents and deduplicate against $\mathcal{P}_g$ and within the batch
            \State Form $\mathcal{P}_{g+1}$ by appending children to $\mathcal{P}_g$ in fixed parent-slot and proposal order, stopping at size $N$
            \State $g\gets g+1$
        \EndWhile
        \State Freeze $\mathcal{P}\gets\mathcal{P}_g$
        \Statex \textbf{Stage II: Validation-based selection}
        \State $\mathcal{O}_0\gets\varnothing$; $\mathcal{H}_0\gets\varnothing$
        \For{$t=0,\ldots,B-1$}
            \State $\mathcal{U}_t\gets\mathcal{P}\setminus\mathcal{O}_t$
            \State $q_{t+1}\sim\pi(\,\cdot\mid\mathcal{U}_t,\mathcal{H}_t)$
            \State Measure $V(q_{t+1})$ on all of $\mathcal{D}_{\mathrm{val}}$
            \State Update $\mathcal{O}_{t+1},\mathcal{H}_{t+1}$ using Eq.~\eqref{eq:selection_policy}
        \EndFor
        \State \Return $p^{\star}=\arg\max_{p\in\mathcal{O}_B}V(p)$
        \Statex Break ties by fixed pool order. The default $\pi$ selects $p_0$ first, then samples uniformly without replacement.
    \end{algorithmic}
\end{algorithm}

\subsection{Candidate-Pool Construction}
\label{sec:pool_generation}
\paragraph{Parent evaluation}
This step spreads search across alternative prompt lineages and amortizes parent-answering cost over multiple proposals. Selecting parents before generating branches gives each selected prompt a shared source of feedback, rather than evaluating it separately for every revision.
Let $g=0,1,\ldots$ index generation batches and $\mathcal{P}_g$ denote the pool at the start of batch $g$, initialized as $\mathcal{P}_0=\{p_0\}$. We uniformly sample up to $P$ distinct parents from this pool without consulting full-validation scores. For each selected parent $p$, we sample a feedback set $T_p\subseteq\mathcal{D}_{\mathrm{tr}}$ of $M$ distinct training examples, then generate and save their responses, scores, and feedback. Parent-specific notation below is local to the current batch.

The feedback set is larger than an individual reflection minibatch, providing multiple combinations of observed successes and failures. Increasing $M$ offers more combinations at greater upfront evaluation cost. Branches may reuse the same examples without regenerating parent responses. Implementation Details specifies the feedback-set size, branch count, and minibatch size.

\paragraph{Parallel reflection}
Different feedback groups can expose different weaknesses of the same parent. We therefore generate several revisions concurrently, allowing alternative corrections to be explored without waiting for an earlier branch's screening result.
For each parent $p$, we sample $K$ reflection minibatches $S_{p,1},\ldots,S_{p,K}$ from its evaluated feedback set $T_p$. Each contains $b$ distinct examples and supplies saved responses, scores, and feedback to a separate revision branch. Small groups focus each reflection on a limited set of observations without repeating parent evaluation.

Every group must include an example on which the parent failed to achieve the maximum task score. Otherwise, it offers no observed error for reflection to address. If the entire feedback set contains no such example, we draw a fresh feedback set and evaluate the same parent again before forming groups. Successful responses may accompany failures, giving the reflector examples of behavior to preserve as well as errors to correct. The sampling policy determines their proportions; we do not require a fixed failure--success ratio across tasks.

Questions may recur across groups, but not within a group; repeated ordered groups are excluded for the same parent in the current batch. Thus, a failed response can inform several branches with different companions rather than being assigned to a single disjoint block. Implementation Details specifies the sampling configurations.

Each minibatch drives a separate reflection and prompt revision. Reflection diagnoses the observed errors before revision uses that diagnosis to modify the reusable prompt. The main task-score experiments use one fixed error-analysis instruction across all branches, keeping reflection style constant across groups and exploration-policy comparisons. Following the trace-based reflection setup~\citep{agrawal2025gepa}, let $\tau(p,S_{p,k})$ contain the saved parent responses, scores, and feedback for $S_{p,k}$. We generate reflection text $r_{p,k}$ and a revised child prompt $p'_{p,k}$ as
\begin{equation}
\label{eq:reflect_rewrite}
\begin{aligned}
    r_{p,k} &\sim R(\,\cdot\mid p,\tau(p,S_{p,k})),\\
    p'_{p,k} &\sim G(\,\cdot\mid p,r_{p,k}),
\end{aligned}
\end{equation}
where $R$ and $G$ are the conditional output distributions of the reflection and revision operators, and $\sim$ denotes sampling. Branches can run concurrently across minibatches and parents; within each branch, reflection precedes revision and child evaluation.

\paragraph{Local screening}
We use a low-cost admission test so that every proposed revision need not undergo full validation. The test asks whether a child improves on the feedback that motivated its revision, reserving broader assessment for the later selection stage.
Each child answers the same $b$ examples used for its reflection and is compared with the saved parent scores. Matching examples avoids confounding the comparison with question selection and requires no additional parent responses. Let $\widehat{v}(p;S)$ be the mean task score of prompt $p$ on minibatch $S$, using recorded responses. The local improvement $\Delta_{p,k}$ and qualifying-child set $\mathcal{Q}_p$ are
\begin{equation}
\label{eq:local_gate}
\begin{aligned}
    \Delta_{p,k}&=\widehat{v}(p'_{p,k};S_{p,k})-\widehat{v}(p;S_{p,k}),\\
    \mathcal{Q}_p&=\{p'_{p,k}:1\leq k\leq K,\ \Delta_{p,k}>0\}.
\end{aligned}
\end{equation}
The gate admits strictly improving revisions to the pool; selection later compares candidates on the complete validation set.

\paragraph{Pool admission}
Keeping multiple qualifying children preserves alternative paths for subsequent refinement instead of selecting a single winner from each parent's proposals. Intermediate prompts can thus support useful descendants without first consuming the full-validation budget.
We combine the qualifying children from all selected parents and retain nonduplicate prompts, together with their parent identifiers and generation metadata. After batch $g$ finishes, candidates are added to $\mathcal{P}_g$ in a predefined parent-slot and proposal order to form $\mathcal{P}_{g+1}$, rather than in completion order. Fixed-order admission prevents response latency from determining which candidates occupy the remaining slots. If the final batch exceeds the target size, this order determines which candidates fill the remaining slots. Construction continues until $N$ prompts, including the initial prompt, have been retained; we denote this frozen pool by $\mathcal{P}$. No full-validation scores are queried during this stage, allowing unvalidated intermediate prompts to support further revisions.

\subsection{Validation-Based Selection}
\label{sec:validation_selection}
Once $\mathcal{P}$ is frozen, an exploration policy $\pi$ chooses which candidates receive full validation. Let $t$ count completed selections, $\mathcal{O}_t$ be the set of fully validated candidates, and $\mathcal{H}_t=\{(p,V(p)):p\in\mathcal{O}_t\}$ be the observed history. The remaining candidates form $\mathcal{U}_t=\mathcal{P}\setminus\mathcal{O}_t$. For $t=0,\ldots,B-1$, let $q_{t+1}$ denote the next selected prompt. The common interface is
\begin{equation}
\label{eq:selection_policy}
\begin{aligned}
    q_{t+1}&\sim\pi(\,\cdot\mid\mathcal{U}_t,\mathcal{H}_t),\\
    \mathcal{O}_{t+1}&=\mathcal{O}_t\cup\{q_{t+1}\},\\
    \mathcal{H}_{t+1}&=\mathcal{H}_t\cup\{(q_{t+1},V(q_{t+1}))\},
\end{aligned}
\end{equation}
starting from empty sets $\mathcal{O}_0=\mathcal{H}_0=\varnothing$. The exploration policy has support only on candidates not yet fully validated and may ignore the observed scores. Test scores are never inputs. This interface separates the exploration policy from candidate-pool construction; we evaluate random selection and MLP + similarity.

\paragraph{Random selection}
In the baseline-inclusive configuration, the initial prompt is selected first, $q_1=p_0$. The remaining $B-1$ candidates are sampled uniformly without replacement from $\mathcal{P}\setminus\{p_0\}$, independently of their validation scores.

\paragraph{MLP + similarity}
We also evaluate an MLP trained on prompt embeddings and observed full-validation scores. After random initialization, the policy repeatedly refits the predictor and ranks unvalidated candidates by a weighted combination of predicted score and novelty. Novelty is one minus the maximum cosine similarity to any already validated prompt. Predictions and novelty are separately min--max normalized over the remaining candidates before combination. This policy uses only acquired validation labels and the frozen pool; unlike the baseline-inclusive random policy, it does not force the initial prompt to be selected.

Selection stops when the validation budget is exhausted. The returned prompt $p^{\star}$ is
\begin{equation}
    p^{\star}=\underset{p\in\mathcal{O}_B}{\operatorname{arg\,max}}\;V(p),
\end{equation}
with ties broken by fixed pool order. Exploration policies share the same pool and final selection criterion, differing in which candidates receive validation. The selection stage requires $B|\mathcal{D}_{\mathrm{val}}|$ candidate--example evaluations, with construction and final testing accounted for separately.

\subsection{Inference}
\label{sec:inference}
We fix $p^{\star}$ and model weights throughout testing. For each unseen input $x$, the solver receives $p^{\star}$ through the task's system-instruction template, together with the question and any task-provided context, and generates $f_{p^{\star}}(x)$. No reflection, candidate search, or prompt-tree traversal is performed. The task-specific evaluator scores the answer against the held-out target without feeding test scores back into selection. Decoding settings and three-seed averaging follow the experimental setup.

\section{Experimental Setup}
\label{sec:experimental_setup}

\subsection{Research Questions}
\label{sec:research_questions}
We organize the experiments around three research questions (RQs):
\begin{itemize}
    \item[\textbf{RQ1.}] How do the two-stage framework and its exploration policies compare with the initial solver, prompt optimization, and preference fine-tuning across two models?
    \item[\textbf{RQ2.}] What do the reported construction and full-validation times indicate about the relative costs of our workflow and GEPA?
    \item[\textbf{RQ3.}] How do the retained candidate trees compare with GEPA in mean depth and mean branching factor?
\end{itemize}

\subsection{Datasets}
\begingroup
\emergencystretch=1em
We evaluate on five benchmarks. \textbf{HotpotQA} \citep{yang2018hotpotqa} tests multi-hop question answering, scored by normalized exact match using three retrieved passages per question. \textbf{SuperGPQA} \citep{map2025supergpqa} contains graduate-level questions with ten answer options, scored by accuracy. \textbf{LiveBench-Math} \citep{white2024livebench} covers mathematical reasoning and uses task-specific answer checkers. \textbf{IFBench} \citep{pyatkin2025ifbench} measures instruction following using the mean fraction of satisfied constraints per example. \textbf{BIG-Bench Extra Hard (BBEH)} \citep{kazemi2025bbeh} evaluates challenging reasoning tasks with binary answer matching.
\par\endgroup

Training, validation, and test subsets contain 150, 300, and 300 examples, respectively, except for LiveBench-Math, which uses 122 examples per subset. For IFBench, training and validation use multi-constraint instruction data, while testing uses the official test set. All subsets are fixed across methods; validation guides prompt selection, and test data are reserved for final evaluation.

\subsection{Models}
We evaluate \textbf{Qwen3.5-9B} \citep{qwen2026qwen35} and \textbf{Llama-3.1-8B-Instruct} \citep{grattafiori2024llama3}. Within each prompt-optimization experiment, the same model performs task solving, reflection, and prompt revision, with its parameters fixed; Qwen's thinking mode is disabled. Final task inference uses temperature $0.7$, as do reflection and revision. Task responses have a maximum output budget of 4,096 tokens. For every method, final test scores are averaged over three inference seeds.

\subsection{Compared Methods}
We compare against four reference methods: the initial solver, MIPRO version 2 (MIPROv2) \citep{opsahl-ong-etal-2024-optimizing}, GEPA \citep{agrawal2025gepa}, and Direct Preference Optimization (DPO) \citep{rafailov2023dpo}. They cover an unoptimized prompt, prompt optimization, and preference-based parameter adaptation. Within each model block, methods start from the same base checkpoint and use the same test examples and scoring metrics.

\paragraph{Initial solver}
The initial solver receives only the original question and the required output format, without an optimized prompt or reflective search.

\paragraph{MIPROv2}
MIPROv2 extends MIPRO \citep{opsahl-ong-etal-2024-optimizing} in the Declarative Self-improving Python (DSPy) framework \citep{khattab2024dspy} to jointly optimize instructions and few-shot demonstrations.\footnote{\url{https://dspy.ai/3.4.0/api/optimizers/MIPROv2/}} It first constructs demonstration candidates from successful executions and proposes instructions informed by the task, program, and examples. Bayesian optimization then searches over combinations of instructions and demonstrations using validation feedback. The resulting program retains the model's parameters and improves its behavior through the selected prompt configuration.

\paragraph{GEPA}
GEPA \citep{agrawal2025gepa} is a reflective prompt optimizer that combines natural-language feedback with Pareto-based candidate selection. It examines execution trajectories to diagnose errors and propose revised instructions, then evaluates candidates to guide further search. By retaining prompts that perform well on different examples, it preserves complementary strengths rather than relying only on a single aggregate-score winner. Reflection and selection are interleaved throughout optimization, improving task prompts without updating the underlying model's parameters.
% Existing five-task GEPA configurations all specify max_full_evaluations=30.
% 20 rounds of GEPA runtime results were supplied separately by the author on 2026-10-19.

\paragraph{DPO}
DPO \citep{rafailov2023dpo} is a preference-based method for adapting language models. Given preferred and dispreferred responses to the same input, it optimizes a pairwise objective that increases the relative likelihood of preferred responses with respect to a fixed reference policy. This formulation avoids training a separate reward model or running a reinforcement-learning optimization loop. Unlike prompt optimization, DPO changes trainable model parameters to incorporate preference information rather than searching over natural-language instructions.

\paragraph{Our exploration policies}
Both variants use a single fixed reflection instruction and the same frozen candidate pool for each model--dataset pair. \textit{Ours (Random)} uses baseline-inclusive random selection; \textit{Ours (MLP + similarity)} combines predicted validation performance with similarity-based exploration, as described in Section~\ref{sec:validation_selection}.

% Place the double-column float before the final setup subsection so that
% it appears above the results discussion rather than among the references.
% Main scores supplied by the author on 2026-10-19 at 11:25:59.
% This update supersedes the 2026-10-13 scores, including the former
% BBEH First-20 comparison. Do not alter historical experiment artifacts.
% IO is displayed consistently as Initial solver; GEPA-30 as GEPA,
% with the full-validation budget specified in the setup.
% Both ours rows use one ordinary reflector, not one proposal branch.
% Report scores/means only, without adding standard deviations.
\begin{table*}[t]
    \caption{\textbf{Main task scores (\%; higher is better).} Scores are three-seed test means. Each of our selection policies uses a budget of 20 full-validation evaluations, compared with 30 for GEPA. Average is the unweighted mean across five benchmarks. Bold and underline denote column-wise best and second-best scores within each model block.}
    \label{tab:main_results}
    \centering
    \small
    \setlength{\tabcolsep}{4pt}
    \begin{tabular*}{\textwidth}{@{\extracolsep{\fill}}lcccccc@{}}
        \toprule
        Method & SuperGPQA & LiveBench-Math & IFBench & BBEH & HotpotQA & Average \\
        \midrule
        \multicolumn{7}{c}{\textbf{Llama-3.1-8B-Instruct}} \\
        \midrule
        Initial solver & $17.00$ & $7.38$ & $31.00$ & $9.33$ & $32.33$ & $19.41$ \\
        MIPROv2 & $18.67$ & $8.47$ & $31.50$ & $9.33$ & $\mathbf{44.89}$ & $\underline{22.57}$ \\
        GEPA & $16.67$ & $\mathbf{10.66}$ & $30.78$ & $7.33$ & $41.00$ & $21.29$ \\
        DPO & $16.33$ & $7.38$ & $\mathbf{32.00}$ & $8.33$ & $41.00$ & $21.01$ \\
        Ours (Random) & $\underline{18.89}$ & $\underline{10.38}$ & $28.11$ & $\mathbf{10.00}$ & $\underline{42.44}$ & $21.96$ \\
        Ours (MLP + similarity) & $\mathbf{21.44}$ & $9.02$ & $\underline{31.67}$ & $\underline{9.67}$ & $42.22$ & $\mathbf{22.80}$ \\
        \midrule
        \multicolumn{7}{c}{\textbf{Qwen3.5-9B}} \\
        \midrule
        Initial solver & $48.44$ & $53.28$ & $39.28$ & $\underline{25.22}$ & $40.67$ & $41.38$ \\
        MIPROv2 & $50.00$ & $51.64$ & $31.33$ & $22.00$ & $49.33$ & $40.86$ \\
        GEPA & $50.56$ & $52.05$ & $40.39$ & $21.11$ & $51.00$ & $43.02$ \\
        DPO & $49.33$ & $52.46$ & $36.83$ & $21.33$ & $40.00$ & $39.99$ \\
        Ours (Random) & $\underline{51.33}$ & $\underline{56.28}$ & $\mathbf{41.67}$ & $24.56$ & $\mathbf{51.89}$ & $\underline{45.15}$ \\
        Ours (MLP + similarity) & $\mathbf{52.78}$ & $\mathbf{56.83}$ & $\underline{41.22}$ & $\mathbf{25.78}$ & $\underline{51.78}$ & $\mathbf{45.68}$ \\
        \bottomrule
    \end{tabular*}
\end{table*}

\subsection{Implementation Details}
\label{sec:implementation_details}
Experiments use NVIDIA A100 GPUs with 40\,GB or 80\,GB of memory per GPU. We report candidate construction and full validation separately in minutes, with 20 full-validation evaluations per method in the runtime comparison. The timings describe the reported execution settings; construction and validation are measured as separate stages.

\paragraph{Prompt optimization}
We set the target pool size to $N=100$ prompts, including the initial prompt, using the training-only procedure in Section~\ref{sec:pool_generation}. Each generation batch samples up to $P=5$ parents. Each parent first answers $M=15$ training examples. From its scored responses, we sample $K=5$ minibatches of $b=3$ examples. Across all benchmarks, each minibatch contains distinct examples and at least one parent error; examples may recur across minibatches, but repeated ordered minibatches are rejected. Each child is screened on its corresponding three examples against saved parent scores, not on the full feedback set. All branches use the same ordinary error-analysis instruction.

After freezing the pool, the random policy validates the initial prompt and 19 additional candidates sampled uniformly without replacement. The MLP + similarity policy uses five randomly selected candidates for initialization, then allocates the remaining 15 validations sequentially. It embeds prompt text with a frozen MPNet-based sentence encoder \citep{song2020mpnet} and refits an MLP with one 16-unit hidden layer, tanh activation, an L2 regularization coefficient of $1$, and a maximum of 500 quasi-Newton optimization iterations. Acquisition weights are $0.4$ for normalized predicted score and $0.6$ for normalized novelty. Both policies select the prompt with the highest measured validation score, breaking ties by fixed pool order. For task-score comparisons, our method uses 20 full-validation evaluations, and GEPA is configured with a budget of 30.

\paragraph{DPO training}
For each training question, we sample four answers at temperature $0.7$ with a maximum output length of 4,096 tokens. LoRA uses rank $16$, scaling parameter $\alpha=32$, and dropout $0.05$. We train for two epochs with learning rate $5\times10^{-6}$, four gradient-accumulation steps, and DPO preference-loss coefficient $\beta=0.1$. We use the second-epoch checkpoint throughout. Testing uses the fixed test examples and native scorers, with scores averaged over three inference seeds.
% The author updated the test protocol to three inference seeds for all methods.

\section{Results}
\subsection{Main Results (RQ1)}
\label{sec:main_results}

Table~\ref{tab:main_results} shows that our framework achieves the highest reported score in most model--dataset combinations. The two exploration policies collectively lead in seven of ten comparisons, covering all five Qwen tasks as well as Llama SuperGPQA and BBEH. MLP + similarity leads on Qwen SuperGPQA, LiveBench-Math, and BBEH, and on Llama SuperGPQA; random selection leads on Qwen IFBench and HotpotQA, and on Llama BBEH. Neither selector consistently leads across tasks.

The improvements over the initial solver extend across both model families and are largest on HotpotQA. Random selection increases Qwen from 40.67 to 51.89 and Llama from 32.33 to 42.44, gains of 11.22 and 10.11 percentage points. MLP + similarity reaches 51.78 and 42.22, respectively, yielding similarly large gains. On SuperGPQA, this policy increases Qwen from 48.44 to 52.78 and Llama from 17.00 to 21.44, improvements of 4.34 and 4.44 points despite different starting scores. More broadly, it improves over the initial solver in all ten comparisons, while random selection improves in eight. Across the same comparisons, MIPROv2 improves in six cases, GEPA in five, and DPO in three. The gains therefore extend beyond a few favorable tasks.

Our method also outperforms GEPA in most comparisons. Random selection and MLP + similarity exceed GEPA in eight and nine of ten cases, respectively, with both policies winning on all five Qwen tasks. Their largest Qwen margins occur on LiveBench-Math, where 56.28 and 56.83 exceed GEPA's 52.05 by 4.23 and 4.78 points. On Llama SuperGPQA, MLP + similarity reaches 21.44 versus 16.67, a 4.77-point gain; random selection reaches 10.00 versus 7.33 on Llama BBEH. On Llama LiveBench-Math, GEPA's 10.66 exceeds random selection at 10.38 and MLP + similarity at 9.02. On Llama IFBench, random selection scores 28.11 versus GEPA's 30.78, whereas MLP + similarity reaches 31.67.

The remaining gaps to the best baseline are concentrated on Llama. GEPA leads on LiveBench-Math at 10.66, DPO on IFBench at 32.00, and MIPROv2 on HotpotQA at 44.89; our best corresponding scores are 10.38, 31.67, and 42.44. In contrast, both policies exceed MIPROv2 and DPO on every Qwen benchmark. This model-dependent pattern may reflect differences in reflection quality or the ability to follow revised instructions, since each checkpoint serves as both solver and reflector. Recent prompt-evolution studies, including GEPA and C-Evolve, evaluate Qwen3-8B alongside other models \citep{agrawal2025gepa,li2025cevolve}. Our results extend this comparison to Llama, capturing the combined effect of each model's task-solving and reflective-revision capabilities.

Prompt optimization is competitive with preference fine-tuning without updating solver weights. Each of our policies exceeds DPO in nine of ten comparisons; Llama IFBench is the sole exception. On Qwen LiveBench-Math, random selection and MLP + similarity score 56.28 and 56.83 against DPO's 52.46, while on HotpotQA they score 51.89 and 51.78 against 40.00. These results compare final performance under the evaluated configurations, not equal-cost optimization.

Random selection nevertheless exhibits two regressions that illustrate the importance of candidate choice. On Qwen BBEH, it reaches 24.56, above GEPA's 21.11 but below the initial solver's 25.22. MLP + similarity reaches 25.78 and improves upon both. On Llama IFBench, random selection falls from the initial score of 31.00 to 28.11, whereas MLP + similarity reaches 31.67. In contrast, random selection leads on Qwen IFBench at 41.67 versus 41.22 for the MLP + similarity policy. It avoids both regressions, while the contrast between the two IFBench results shows that selector performance depends on the model.

MLP + similarity beats random selection in five comparisons and loses in five. Its largest gains are 3.56 points on Llama IFBench and 2.55 on Llama SuperGPQA, while random selection leads by 1.36 points on Llama LiveBench-Math. Since both policies use the same pool, their shared gains are consistent with parallel revision and local screening contributing useful candidates; the selector changes validation allocation rather than construction.

MLP + similarity reaches 22.80 on Llama and 45.68 on Qwen, compared with the strongest baseline averages of 22.57 for MIPROv2 and 43.02 for GEPA. Random selection averages 21.96 and 45.15, exceeding GEPA's 21.29 and 43.02 without MLP + similarity.

The validation budget also determines the number of scored examples. On benchmarks with 300 validation examples, our 20 full validations require 6,000 evaluations, compared with 9,000 for GEPA's 30. On LiveBench-Math, the counts are 2,440 and 3,660. This one-third reduction concerns selection; construction costs are reported separately below. Despite the smaller validation budget, MLP + similarity improves the model-level average over GEPA by 1.51 points on Llama and 2.66 on Qwen; random selection gains 0.67 and 2.13 points, respectively.

\subsection{Runtime Evaluation (RQ2)}
\label{sec:runtime}

% Stage-wise timings updated from author screenshot 2026-10-19 at 17:43:13.
% Units: minutes. Only Qwen IFBench construction was recomputed:
% 37.02 seconds * 3.5 / 6.24 * 100 / 60 = 34.607... minutes.
% The screenshot states that the other cells have not been uniformly recomputed.
% Author correction, 2026-10-26: the reported 34.61 minutes is measured.
% The underlying timing log was not independently checked for this edit.
\begin{table}[ht]
    \caption{\textbf{Reference runtime in minutes.} Times are intended for relative comparison within our experimental setting. Build denotes candidate construction.}
    \label{tab:runtime}
    \centering
    \small
    \setlength{\tabcolsep}{2pt}
    \begin{tabular*}{\columnwidth}{@{\extracolsep{\fill}}lrrrr@{}}
        \toprule
        & \multicolumn{2}{c}{Ours (Random)} & \multicolumn{2}{c}{20 rounds of GEPA} \\
        \cmidrule(lr){2-3}\cmidrule(l){4-5}
        Dataset & Build & Validation & Build & Validation \\
        \midrule
        \multicolumn{5}{c}{\textbf{Llama-3.1-8B-Instruct}} \\
        \midrule
        SuperGPQA & 53.92 & 54.42 & 85.91 & 56.58 \\
        LiveBench-Math & 219.51 & 35.54 & 287.26 & 37.24 \\
        IFBench & 38.77 & 43.33 & 46.02 & 40.97 \\
        BBEH & 173.17 & 76.37 & 367.25 & 78.51 \\
        HotpotQA & 15.70 & 15.28 & 25.29 & 19.42 \\
        \midrule
        \multicolumn{5}{c}{\textbf{Qwen3.5-9B}} \\
        \midrule
        SuperGPQA & 87.47 & 52.04 & 63.70 & 64.01 \\
        LiveBench-Math & 113.12 & 47.58 & 113.80 & 45.98 \\
        IFBench & 34.61 & 46.92 & 33.89 & 52.33 \\
        BBEH & 284.64 & 100.34 & 358.66 & 63.17 \\
        HotpotQA & 33.34 & 8.96 & 51.75 & 8.98 \\
        \bottomrule
    \end{tabular*}
\end{table}

Table~\ref{tab:runtime} reports construction and validation times for our method and 20 rounds of GEPA. Both methods use 20 full validations here, whereas the task-score comparison gives GEPA 30. Combined totals are lower for our method in eight of ten comparisons: all five Llama tasks and Qwen IFBench, BBEH, and HotpotQA. For Qwen IFBench, the totals are $34.61+46.92=81.53$ versus $33.89+52.33=86.22$ minutes. Qwen SuperGPQA takes longer at 139.51 versus 127.71 minutes, while LiveBench-Math is nearly tied at 160.70 versus 159.78. These reference timings reflect the reported execution settings; hardware-normalized speedups require matched conditions.

The breakdown locates the main differences. On Llama BBEH, construction accounts for 194.08 minutes of the 196.22-minute total difference. On Qwen BBEH, shorter construction saves 74.02 minutes but longer validation offsets 37.17 minutes, leaving a net difference of 36.85 minutes. Thus, the construction advantage can be partly offset at selection, even with the same full-validation count.

\subsection{Search-Tree Structure (RQ3)}
\label{sec:tree_depth}
The third experiment characterizes the retained search trees using mean depth and mean branching factor. Depth counts revision steps from the initial prompt; its zero depth is included in the mean over all retained nodes. Mean branching factor is the number of retained parent--child edges divided by the number of non-leaf nodes, measuring the average number of direct children per expanded node. Table~\ref{tab:tree_depth} reports both statistics.

% Mean depth and branching factor: author screenshot 2026-10-19 at 4.40.31 PM.
% Branching factor = retained parent-child edges / non-leaf nodes.
% Mean depth includes the initial prompt at depth zero.
% Interpretation and snapshot-size caveats: companion screenshot at 4.40.34 PM.
\begin{table}[ht]
    \caption{\textbf{Retained search-tree structure.} Depth and branching describe search behavior, not prompt quality. Node counts differ across methods.}
    \label{tab:tree_depth}
    \centering
    \small
    \setlength{\tabcolsep}{2pt}
    \begin{tabular*}{\columnwidth}{@{\extracolsep{\fill}}lrrrr@{}}
        \toprule
        & \multicolumn{2}{c}{Mean depth} & \multicolumn{2}{c}{Mean branching} \\
        \cmidrule(lr){2-3}\cmidrule(l){4-5}
        Dataset & Ours & GEPA & Ours & GEPA \\
        \midrule
        \multicolumn{5}{c}{\textbf{Llama-3.1-8B-Instruct}} \\
        \midrule
        SuperGPQA & 4.37 & 3.47 & 2.41 & 1.81 \\
        LiveBench-Math & 4.09 & 3.00 & 2.15 & 2.23 \\
        IFBench & 3.37 & 5.73 & 3.19 & 1.93 \\
        BBEH & 5.24 & 3.80 & 2.11 & 1.93 \\
        HotpotQA & 5.01 & 3.67 & 2.75 & 2.07 \\
        \midrule
        \multicolumn{5}{c}{\textbf{Qwen3.5-9B}} \\
        \midrule
        SuperGPQA & 3.32 & 2.73 & 2.68 & 2.08 \\
        LiveBench-Math & 3.44 & 1.67 & 2.83 & 2.20 \\
        IFBench & 3.28 & 2.93 & 2.54 & 1.81 \\
        BBEH & 3.46 & 3.00 & 2.36 & 2.42 \\
        HotpotQA & 3.28 & 3.10 & 2.75 & 2.07 \\
        \bottomrule
    \end{tabular*}
\end{table}

Under their respective pool sizes and retention rules, our searches produce trees with greater mean branching in eight of ten comparisons, with four of five tasks for each model; the exceptions are Llama LiveBench-Math and Qwen BBEH. Mean depth is greater in nine of ten comparisons, with Llama IFBench the exception at 3.37 versus 5.73. Expanded nodes therefore generally retain more alternatives, and revision paths extend through more generations.

Depth and branching need not move together. On Llama IFBench, higher branching of 3.19 versus 1.93 accompanies lower depth of 3.37 versus 5.73, indicating more direct alternatives per expanded node rather than longer chains. Llama LiveBench-Math shows the reverse: depth rises from 3.00 to 4.09 while branching falls from 2.23 to 2.15. Four Qwen tasks gain in both dimensions. Together, these measures distinguish expansion around a parent from refinement along a path.

\section{Conclusion}
\label{sec:conclusion}
We presented a two-stage framework that separates training-driven candidate construction from full-validation-based selection. Shared parent feedback supports parallel reflection, while later descendants can be selected without fully validating intermediate prompts. Across five benchmarks and two models, our variants lead on all five Qwen tasks and two Llama tasks, but neither exploration policy dominates. Reference timing totals are lower in eight of ten comparisons against 20 rounds of GEPA in the reported setting. Retained trees show greater mean depth in nine comparisons and greater branching in eight, under differing node counts and retention rules. Better candidate-selection policies and matched-total-budget comparisons remain important directions.

\section*{Limitations}
MLP + similarity predicts scores from individual prompt embeddings, without modeling reflection paths. Future work could replace the MLP with a recurrent neural network (RNN) that encodes prompts, reflections, and revisions along each parent--child path. The resulting trajectory representations could exploit evolution history to predict candidate performance and guide full-validation selection.

Our experiments cover two open-weight models, Qwen3.5-9B and Llama-3.1-8B-Instruct, but no closed-source models such as GPT \citep{openai2023gpt4} or Claude \citep{anthropic2024claude3}. Whether the results extend to these model families remains to be tested.

\clearpage
\bibliographystyle{ACM-Reference-Format}
\bibliography{references}

\appendix
\section{Generative AI Usage}
We used generative AI tools to assist with manuscript writing and code development. The authors take responsibility for the content of the paper and the code.

\end{document}